\documentclass[runningheads]{llncs}
\usepackage{graphicx}

\usepackage{tikz}
\usepackage{comment}
\usepackage{amsmath,amssymb} 
\usepackage{color}

\usepackage[ruled]{algorithm2e}
\usepackage{subfigure}
\usepackage{multirow}
\usepackage{booktabs}

\begin{document}
\pagestyle{headings}
\mainmatter
\def\ECCVSubNumber{1789}  

\title{Anti-Bandit Neural Architecture Search for Model Defense} 

\titlerunning{ABanditNAS}
%
\author{Hanlin Chen\inst{1} \and
Baochang Zhang\inst{1}\thanks{Corresponding author.} \and
Song Xue\inst{1} \and Xuan Gong\inst{2*} \and Hong Liu\inst{3} \and Rongrong Ji\inst{3} \and David Doermann\inst{2}}

%
\authorrunning{H. Chen et al.}
%
\institute{Beihang University, Beijing, China \and University at Buffalo, Buffalo, USA
\and Xiamen University, Fujian, China, \\
\email{\{hlchen,bczhang\}@buaa.edu.cn}}
\maketitle

\begin{abstract}
Deep convolutional neural networks (DCNNs) have dominated as the best performers in machine learning, but can be challenged by adversarial attacks. In this paper, we defend against adversarial attacks using neural architecture search (NAS) which is based on a comprehensive search of denoising blocks, weight-free operations, Gabor filters and convolutions.  The resulting anti-bandit NAS (ABanditNAS) incorporates a new operation evaluation measure and search process based on the lower and upper confidence bounds (LCB and UCB). Unlike the conventional bandit algorithm using UCB for evaluation only, we use  UCB to abandon arms for search efficiency and LCB for a fair competition between arms. Extensive experiments demonstrate that ABanditNAS is faster than other NAS methods, while achieving an $8.73\%$ improvement over prior arts on CIFAR-10 under PGD-$7$ \footnote{Some test models are available in 
\url{https://github.com/bczhangbczhang/ABanditNAS}.}.
\end{abstract}

\section{Introduction}

The success of deep learning models \cite{bengio2017deep}  have been demonstrated on various computer vision tasks such as image classification \cite{he2016deep}, instance segmentation \cite{long2015fully} and object detection \cite{szegedy2015going}. However, existing deep models are sensitive to adversarial attacks \cite{carlini2017towards,goodfellow2014explaining,szegedy2013intriguing}, where adding an imperceptible perturbation to input images can cause the models to perform incorrectly. Szegedy et. al \cite{szegedy2013intriguing} also observe that these adversarial examples are transferable across multiple models such that adversarial examples generated for one model might mislead other models as well. Therefore, models deployed in the real world scenarios are susceptible to adversarial attacks \cite{liu2016delving}. While many methods have been proposed to defend against these attacks \cite{szegedy2013intriguing,cubuk2017intriguing}, improving the network training process proves to be one of the most popular.  These methods inject adversarial examples into the training data to retrain the network \cite{goodfellow2014explaining,kurakin2016adversarial,athalye2018obfuscated}. Similarly, pre-processing defense methods modify adversarial inputs to resemble clean inputs \cite{samangouei2018defensegan,liao2018defense}  by transforming the adversarial images into clean images before they are fed into the classifier.  

Overall, however, finding adversarially robust architectures using neural architecture search (NAS) shows even more promising results \cite{cubuk2017intriguing,vargas2019evolving,guo2019meets,dong2019neural}. NAS has attracted a great attention with remarkable performance in various deep learning tasks. In \cite{cubuk2017intriguing} the researchers investigate the dependence of adversarial robustness on the network architecture via NAS. A neural architecture search framework for adversarial medical image segmentation is proposed by \cite{dong2019neural}. \cite{guo2019meets} leverages one-shot NAS \cite{BenderKZVL18} to understand the influence of network architectures against adversarial attacks.  Although promising performance is achieved in existing NAS based methods, this direction still remains largely unexplored.

	\begin{figure*}[t]
		\centering
		\includegraphics[scale=.5]{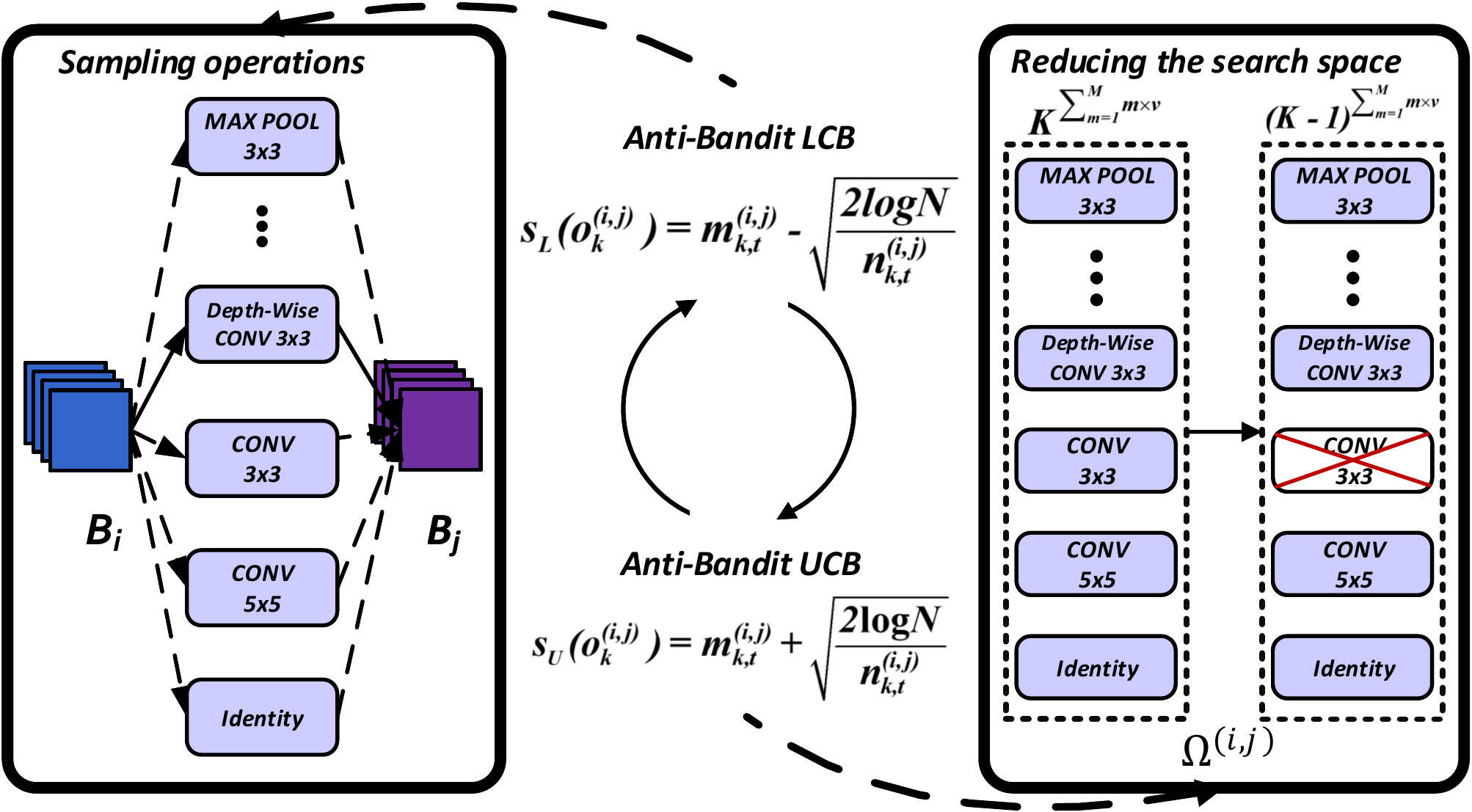}
		\caption{ABanditNAS is mainly divided into two steps: sampling using LCB and abandoning based on UCB.}
		\label{fig:anti-bandit}
	\end{figure*}

In this paper, we consider NAS for model defense by treating it as a multi-armed bandit problem and introduce a new anti-bandit algorithm into adversarially robust network architecture search. To improve the robustness to adversarial attacks, a comprehensive search space is designed by including diverse operations, such as denoising blocks, weight-free operations, Gabor filters and convolutions. However, searching a robust network architecture  is  more challenging than traditional NAS, due to the complicated search space, and learning inefficiency caused by adversarial training. We develop an anti-bandit algorithm based on both the upper confidence bound (UCB)  and the lower confidence bound (LCB) to handle the  huge and complicated search space, where the number of operations that define the space can be $9^{60}$! Our anti-bandit algorithm uses UCB  to reduce the search space, and LCB  to   guarantee that every arm is fairly tested before being abandoned.
	
 Making use of the LCB, operations which have poor performance early, such as parameterized operations, will be given more chances  but they are thrown away once they are confirmed to be bad. Meanwhile,  weight-free operations will be  compared with  parameterized operations only when they are well trained.  Based on the observation that the early optimal operation is not necessarily the optimal one in the end, and the worst operations in the early stage usually has a worse performance at the end \cite{zheng2019dynamic}, we exploit UCB to prune the worst operations earlier, after a fair performance evaluation via  LCB. This means that the operations we finally reserve are certainly a near optimal solution. On the other hand, with the operation pruning process, the search space becomes smaller and smaller, leading to an efficient search process. Our framework shown in Fig. \ref{fig:anti-bandit} highlights the anti-bandit NAS (ABanditNAS) for finding a robust architecture from a very complicated search space. The contributions of our paper are as follows:
	\begin{itemize}
		\item
		ABanditNAS is developed to solve the adversarially robust optimization and architecture search in a unified framework. We introduce an anti-bandit algorithm based on a specific   operation search strategy with a lower and an upper bound, which can learn a robust architecture based on a comprehensive operation space.
		\item
		The search space is greatly reduced by our anti-bandit pruning method which abandons operations with less potential, and significantly reduces the search complexity from exponential to polynomial, i.e., $\mathcal{O}(K^{|\mathcal{E_M}| \times v})$ to $\mathcal{O}(K^2 \times T)$ (see Section $3.4$ for details).
		\item
		Extensive experiments demonstrate that the proposed algorithm achieves  better performance than other adversarially robust models on commonly used MNIST and CIFAR-10.
	\end{itemize}

\section{Related Work}

{\bfseries Neural architecture search (NAS).} NAS becomes one of the most promising technologies in the  deep learning paradigm. Reinforcement learning (RL) based methods \cite{zoph2018learning,zoph2016neural} train and evaluate more than $20,000$ neural networks across $500$ GPUs over $4$ days. The recent  differentiable architecture search (DARTS) reduces the search time by formulating the task in a differentiable manner \cite{liu2018darts}.  However, DARTS and its variants \cite{liu2018darts,Xu2019PC} might be less efficient for a complicated search space. To speed up the search process, a \emph{one-shot} strategy is introduced  to do NAS within a few GPU days \cite{liu2018darts,pham2018efficient}. In this one-shot architecture search, each architecture in the search space is considered as a sub-graph sampled from a super-graph, and the search process can be accelerated by parameter sharing \cite{pham2018efficient}. Though \cite{cubuk2017intriguing} uses NAS with reinforcement learning to find adversarially robust architectures that achieve good results, it is insignificant compared to the search time. Those methods also seldom consider high diversity in operations  closely related to model defense in the search strategy.

{\bfseries Adversarial attacks.}  Recent research has shown that neural networks exhibit significant vulnerability to adversarial examples. After the discovery of adversarial examples by \cite{szegedy2013intriguing}, \cite{goodfellow2014explaining} proposes the Fast Gradient Sign Method (FGSM) to generate adversarial examples with a single gradient step. Later, in \cite{kurakin2016adversarial}, the researchers propose the Basic Iterative Method (BIM),  which takes multiple and smaller FGSM steps to improve FGSM, but  renders the adversarial training very slow. This iterative adversarial attack is further strengthened by adding multiple random restarts, and is also incorporated into the adversarial training procedure. In addition, projected gradient descent (PGD) \cite{madry2017towards} adversary attack, a variant of BIM with a uniform random noise as initialization, is recognized to be one of the most powerful first-order attacks \cite{athalye2018obfuscated}. Other popular attacks include the Carlini and Wagner Attack \cite{carlini2017towards} and Momentum Iterative Attack \cite{dong2018boosting}. Among them, \cite{carlini2017towards} devises state-of-the-art attacks under various pixel-space $l_{p}$ norm-ball constraints by proposing multiple adversarial loss functions.

       {\bfseries Model defense.} In order to resist attacks, various methods have been proposed. A category of defense methods improve network’s training regime to counter adversarial attacks. The most common method is adversarial training \cite{kurakin2016adversarial,na2017cascade} with adversarial examples added to the training data. In \cite{madry2017towards},  a defense method called Min-Max optimization is introduced to augment the training data with first-order attack samples. \cite{wang2019bilateral} investigates fast training of adversarially robust models to perturb both the images and the labels during training.
 There are also some model defense methods that target at removing adversarial perturbation by transforming the input images before feeding them to the  network \cite{liao2018defense,athalye2018obfuscated,gupta2019ciidefence}. In \cite{dziugaite2016study,das2017keeping},  the effect of JPEG compression is investigated for removing adversarial noise.  In \cite{osadchy2017no},  the authors apply a set of filters such as median filters and averaging filters to remove perturbation. In \cite{YangME},  a ME-Net method is introduced  to destroy the adversarial noise and re-enforce the global structure of the original images. With the development of NAS, finding adversarially robust architectures using NAS is another promising direction \cite{cubuk2017intriguing}, which is worth in-depth exploration. Recently, { \cite{dong2019neural}  designs three types of primitive operation set in the search space to automatically find two-cell architectures  for semantic image segmentation, especially medical image segmentation, leading to a NAS-Unet backbone network. }
        
    In this paper, an anti-bandit algorithm is introduced  into NAS, and we develop a new  optimization framework to  generate adversarially robust networks. Unlike \cite{ilyas2018prior} using bandits to produce black-box adversarial samples, we propose an anti-bandit algorithm to obtain a robust network architecture. {In addition, existing NAS-based  model defense methods either target at different applications from ours or are less efficient for object classification \cite{cubuk2017intriguing,vargas2019evolving,guo2019meets,dong2019neural}}.

\section{Anti-Bandit Neural Architecture Search}

	\subsection{Search Space}

\begin{figure}[t]
		\subfigure[]{
		    \centering
			\includegraphics[scale=.4]{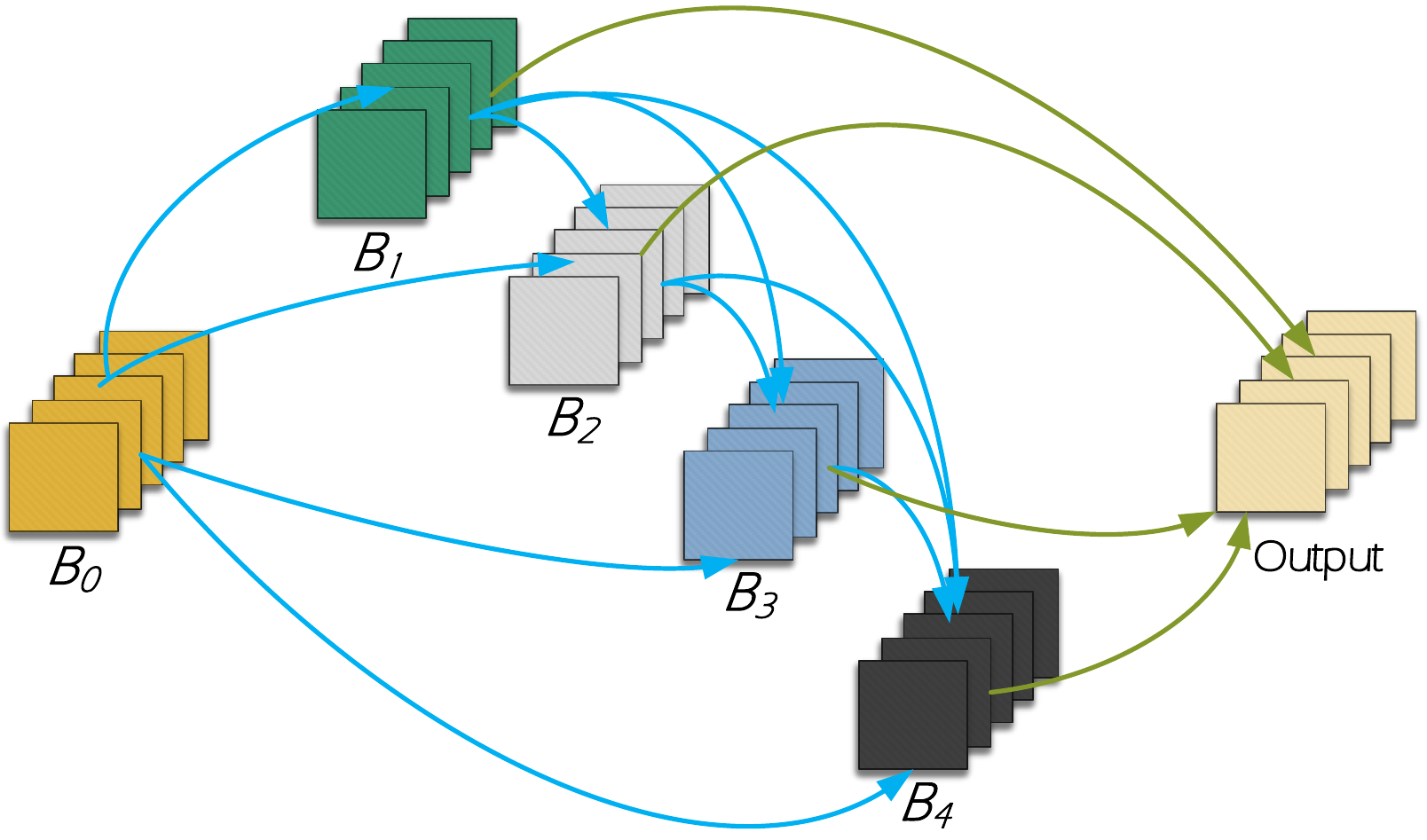}
			\label{fig:cell}
		}
		\hspace{5mm}
		\subfigure[]{
		    \centering
			\includegraphics[scale=.08]{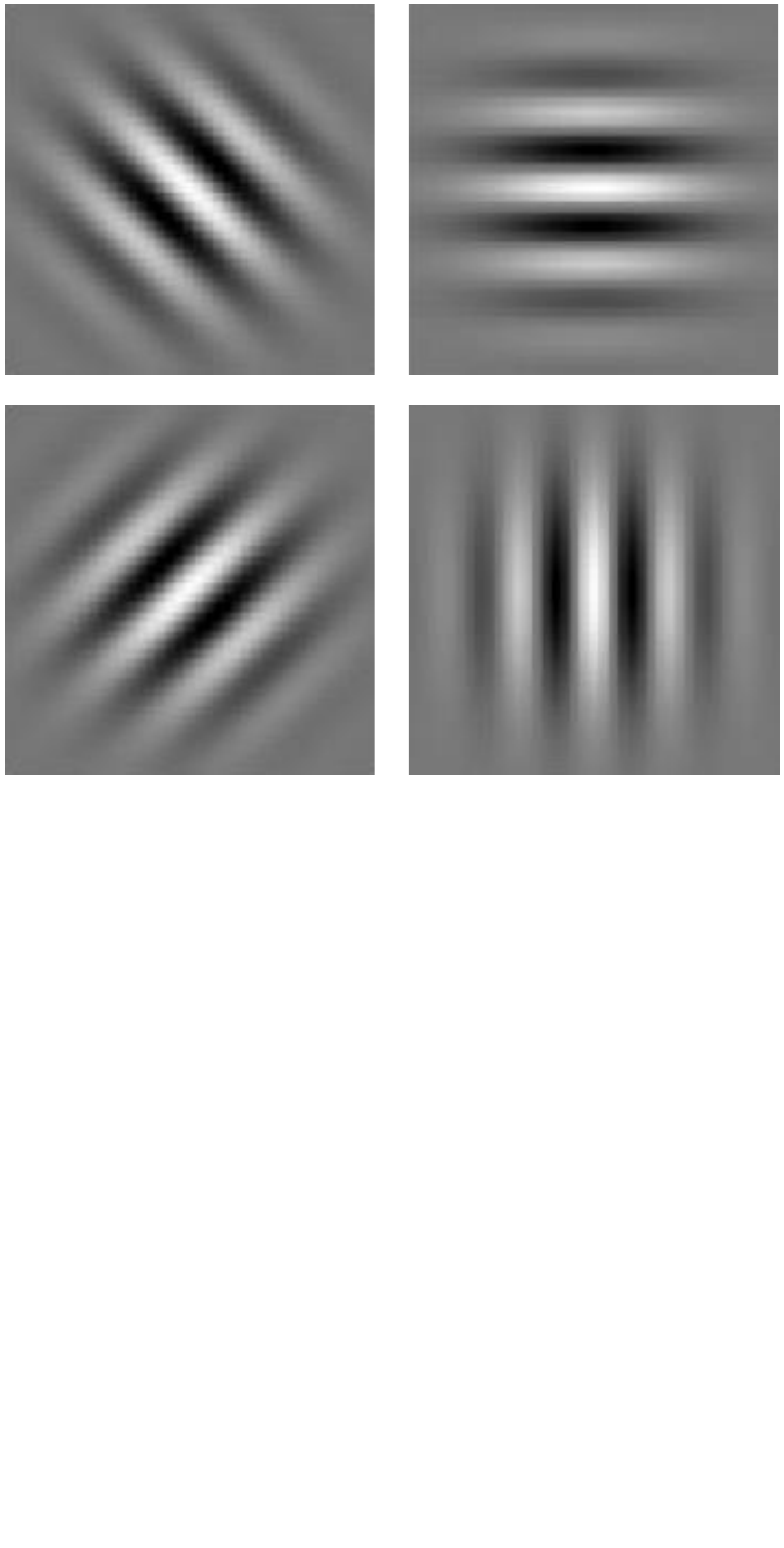}
			\label{fig:gabor}
		}
		\hspace{4mm}
		\subfigure[]{
		    \centering
			\includegraphics[scale=.39]{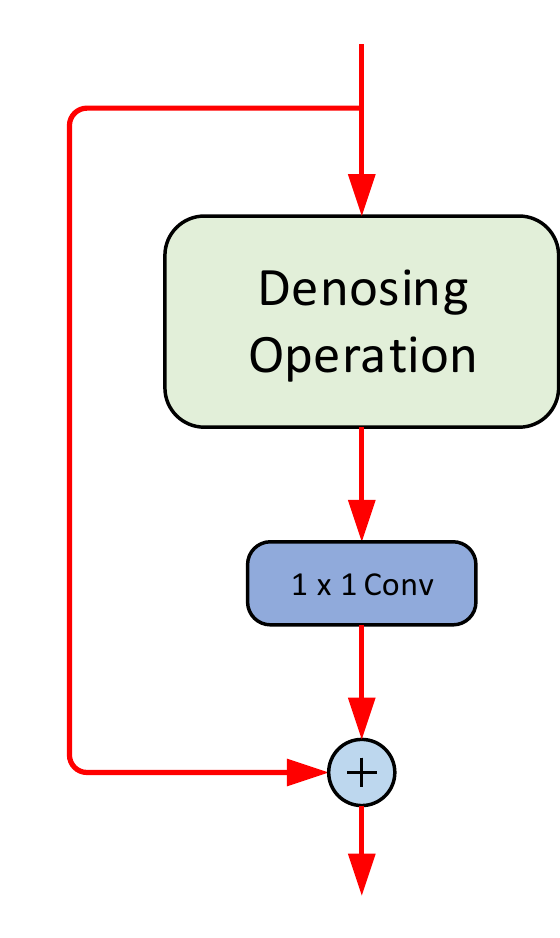}
			\label{fig:denosing}
		}
		\caption{ (a) A cell containing four intermediate nodes $B_1$, $B_2$, $B_3$, $B_4$ that apply sampled operations on the input node $B_0$. $B_0$ is from the output of the previous cell. The output node concatenates the outputs of the four intermediate nodes. (b) Gabor Filter. (c) A generic  denoising block. Following \cite{xie2019feature}, it wraps the denoising operation with a $1\times1$ convolution and an identity skip connection \cite{he2016deep}.}
	\end{figure}

Following \cite{zoph2018learning,liu2018darts,zheng2019dynamic,zhuo2020cp,chen2020bnas}, we search for computation cells as the building blocks of the final architecture. Different from these approaches, we search for $v$ ($v>2$) kinds of cells instead of only normal and reduction cells. Although it increases the search space, our search space reduction in ABanditNAS can make the search efficient enough. A cell is a fully-connected directed acyclic graph (DAG) of $M$ nodes, \emph{i.e.}, $\{B_1, B_2, ..., B_M\}$ as shown in Fig. \ref{fig:cell}. Each node $B_i$ takes its dependent nodes as input, and generates an output through the selected operation $B_j = o^{(i,j)}(B_i).$ Here each node is a specific tensor (\emph{e.g.,} a feature map in convolutional neural networks) and each directed edge $(i,j)$ between $B_i$ and $B_j$ denotes an operation $o^{(i,j)} (.)$, which is sampled from $\Omega^{(i,j)}=\{o^{(i,j)}_1, ..., o^{(i,j)}_K\}$. Note that the constraint $i<j$ ensures that there are no cycles in a cell. Each cell takes the output of the previous cell as input, and we define this node belonging to the previous cell as the input node $B_{0}$ of the current cell for easy description. The set of the operations $\Omega$ consists of $K = 9$ operations. Following \cite{liu2018darts}, there are seven normal operations that are the $3\times3$ max pooling, $3\times3$ average pooling, skip connection (identity), $3\times3$ convolution with rate $2$, $5\times5$ convolution with rate $2$, $3\times3$ depth-wise separable convolution, and $5\times5$ depth-wise separable convolution. The other two are  $3\times3$  Gabor filter and denoising block. Therefore, the size of the whole search space is $K^{|\mathcal{E_M}| \times v}$, where $\mathcal{E_M}$ is the set of possible edges with $M$ intermediate nodes in the fully-connected DAG. The search space of a cell is constructed by the operations of all the edges, denoted as $\{\Omega^{(i,j)}\}$. In our case with $M=4$ and $v=6$, together with the input node, the total number of cell structures in the search space is $9^{(1+2+3+4) \times 6} = 9^{10 \times 6}$. 

	{\bfseries Gabor filter.} Gabor wavelets \cite{gabor1946theory,gabor1946electrical} were invented by Dennis Gabor using complex functions to serve as a basis for Fourier transform in information theory applications. The Gabor wavelets (kernels or filters)   in Fig. \ref{fig:gabor} are defined as: $\exp(-\frac{x'^{2} + \gamma^2y'^{2}}{2\sigma^2})\cos(2\pi \frac{x'}{\lambda}+\psi)$, where $x'=x\cos\theta+y\sin\theta$ and $y'=-x\sin\theta+y\cos\theta$. We set $\sigma$, $\gamma$, $\lambda$, $\psi$ and $\theta$ to be learnable parameters. Note that the symbols used here apply only to the Gabor filter and are different from the symbols used in the rest of this paper. An important property of the wavelets is that the product of its standard deviations is minimized in both time and frequency domains. Also, robustness is another important property which we use here \cite{RobustGabor}.
	
	{\bfseries Denoising block.} In \cite{xie2019feature}, the researchers suggest that adversarial perturbations on images can result in noise in the features. Thus, a denoising block Fig. \ref{fig:denosing} is used to improve adversarial robustness via  feature denoising. Similarly, we add the non-local mean denoising block \cite{buades2005non} to the search space to denoise the features. It computes a denoised feature map $z$ of an input feature map $x$ by taking a weighted mean of the features over all spatial locations $\mathcal{L}$ as 
		$z_p = \frac{1}{C(x)} \sum_{\forall q\in \mathcal{L}} f(x_p,x_q) \cdot x_q$, where $f(x_p,x_q)$ is a feature-dependent weighting function and $C(x)$ is a normalization function. {Also,  the denosing block needs huge computations because of the matrix multiplication between features.}

	It is known that adversarial training is more challenging than that of natural training \cite{shafahi2019adversarial}, which adds an additional burden to NAS. For example, adversarial training using the $F$-step PGD attack needs roughly $F + 1$ times more computation. Also, more operations added to the search space are another burden. To solve these problems, we introduce operation space reduction based on the UCB bandit algorithm into NAS, to significantly reduce the cost of GPU hours, leading to our efficient ABanditNAS.


\subsection{Adversarial Optimization for ABanditNAS}

	Adversarial training \cite{madry2017towards} is a method for learning networks so that they are robust to adversarial attacks. Given a network $f_\theta$ parameterized by $\theta$, a dataset $(x_e,y_e)$, a loss function $l$ and a threat model $\Delta$, the learning problem is typically cast as the following optimization problem: $\min_\theta \sum_e \max_{\delta \in \Delta} l\big(f_\theta (x_e + \delta), y_e \big)$, where $\delta$ is the adversarial perturbation.  A typical choice for a threat model is to take $\Delta = \{ \delta: \| \delta \|_{\infty} \leq \epsilon \}$ for some $\epsilon > 0$, where $\| \cdot \|_{\infty}$ is some $l_\infty$-norm distance metric and $\epsilon$ is the adversarial manipulation budget. This is the $l_\infty$ threat model used by \cite{madry2017towards} and what we consider in this paper. The procedure for adversarial training is to use   attacks to approximate the inner maximization over $\Delta$, followed by some variation of gradient descent on the model parameters $\theta$. For example, one of the earliest versions of adversarial training uses the Fast Gradient Sign Method (FGSM) \cite{goodfellow2014explaining} to approximate the inner maximization. This could be seen as a relatively inaccurate approximation of the inner maximization for $l_\infty$ perturbations, and has the closed form solution: 
		$\theta = \epsilon \cdot \text{sign}\Big(\nabla_x l\big(f(x),y\big)\Big)$.

	A better approximation of the inner maximization is to take multiple, smaller FGSM steps of size $\alpha$ instead. However, the number of gradient computations casused by the multiple steps is proportional to $\mathcal{O}(EF)$ in a single epoch, where $E$ is the size of the dataset and $F$ is the number of steps taken by the PGD adversary. This is $F$ times greater than the standard training which has $\mathcal{O}(E)$ gradient computations per epoch, and so the adversarial training is typically $F$ times slower. To speed up the adversarial training, we combine the FGSM with random initialization \cite{Wong2020Fast}.

	\subsection{Anti-Bandit}
	In machine learning, the multi-armed bandit problem \cite{auer2002finite,alphago} is a classic reinforcement learning (RL) problem that exemplifies the exploration-exploitation trade-off dilemma: shall we stick to an arm that gave high reward so far (exploitation) or rather probe other arms further (exploration)? The Upper Confidence Bound (UCB) is widely used for dealing with the exploration-exploitation dilemma in the multi-armed bandit problem. For example, the idea of bandit is exploited to improve many classical RL methods such as Monte Carlo \cite{lev2006mc} and Q-learning \cite{even2006qlearning}. The most famous one is  AlphaGo \cite{alphago}, which uses the Monte Carlo Tree Search (MCTS) algorithm to play the board game Go but based on a very powerful computing platform unavailable to common researchers.  Briefly, the UCB algorithm chooses at trial the arm $k$ that maximizes 
	\begin{equation}\label{eq:UCB_o}
	    \small
	  	\hat{r}_{k} + \sqrt{\frac{2\log N}{n_{k}}},
	 \end{equation}
	 where $\hat{r}_{k}$ is the average reward obtained from arm $k$, and $n_{k}$ is the number of times arm $k$ has been played up to trial $N$. The first term in Eq.~\ref{eq:UCB_o} is the value term which favors actions that look good historically, and the second is the exploration term which makes actions get an exploration bonus that grows with $\log N$. The total value  can be interpreted as the upper bound of a confidence interval, so that the true mean reward of each arm $k$ with a high probability is below this UCB.

	The  UCB in bandit is not applicable in NAS, because it is  too time-consuming to choose an arm from a huge search space (a huge number of arms, e.g., $9^{60}$), particularly when limited computational resources are available. To solve the problem,  we introduce an anti-bandit algorithm  to reduce the arms for the huge-armed problem by incorporating both the upper confidence bound (UCB) and the lower confidence bound (LCB) into the conventional bandit algorithm. We first define LCB as
	\begin{equation}\label{eq:LCB_o}
	    \small
		\hat{r}_{k} - \sqrt{\frac{2\log N}{n_{k}}}.
	\end{equation}
	 LCB is designed to sample an arm from a huge number of arms for one more trial (later in Eq.~\ref{eq:LCB}). A smaller LCB   means that the less played arm  (a smaller $n_{k}$) is given a bigger chance to be  sampled for a trial.  Unlike  the conventional bandit based on the  maximum UCB (Eq.~\ref{eq:UCB_o}) to choose an arm, our UCB (Eq.~\ref{eq:UCB}) is used to abandon the arm operation with the minimum value, which is why we call our algorithm anti-bandit.
	 
	  Our anti-bandit algorithm is specifically designed for the huge-armed bandit problem by reducing the number of arms based on the UCB. Together with the LCB, it can guarantee  every arm is fairly tested before being abandoned.

	\subsection{Anti-Bandit Strategy for ABanditNAS} \label{Anti-Bandit}
	
	As described in \cite{ying2019bench,zheng2019dynamic}, the validation accuracy ranking of different network architectures is not a reliable indicator of the final architecture quality. However, the experimental results actually suggest a nice property that if an architecture performs poorly in the beginning of training, there is little hope that it can be part of the final optimal model \cite{zheng2019dynamic}. As the training progresses, this observation is more and more certain. Based on this observation, we derive a simple yet effective operation abandoning method. During training, along with the increasing epochs, we progressively abandon the worst performing operation for each edge. Unlike \cite{zheng2019dynamic} which just uses the performance as the evaluation metric to decide which operation should be pruned, we use the anti-bandit algorithm described next to make a decision about which one should be pruned.
		
	\begin{algorithm}[t]
		\small
		\caption{ABanditNAS \label{alg:bandit}}
		\LinesNumbered
		\KwIn{Training data, validation data, searching hyper-graph, adversarial perturbation $\delta$, adversarial manipulation budget $\epsilon$, $K=9$, hyper-parameters $\alpha$, $\lambda=0.7$, $T=3$.} 
		\KwOut{The remaining optimal structure;}
		$t = 0$; $c = 0$; \\
		Get initial performance $m_{k,0}^{(i,j)}$; \\
		\While{$(K>1)$}{
			$c \leftarrow  c + 1$; \\
			$t\leftarrow  t + 1$; \\
			Calculate $s_L(o^{(i,j)}_k)$ using Eq.~\ref{eq:LCB}; \\
			Calculate $p(o^{(i,j)}_k)$ using Eq.~\ref{eq:prob}; \\
			Select an architecture by sampling one operation based on $p(o^{(i,j)}_k)$ from $\Omega^{(i,j)}$ for every edge; \\
			// Train adversarially the selected architecture \\
			\For{$e= 1,...,E$}{
				$\delta$ $=$ Uniform($-\epsilon, \epsilon$); \\
				$\delta \leftarrow \delta + \alpha \cdot $ \text{sign} $\Big(\nabla_x l\big(f(x_e + \delta),y_e\big)\Big)$; \\
				$\delta =\max \big(\min(\delta, \epsilon), -\epsilon\big)$; \\
				$\theta \leftarrow \theta - \nabla_\theta l      \big(f_\theta(x_e + \delta), y_e \big)$; \\
			}
			Get the accuracy $a$ on the validation data; \\
			Update the performance $m_{k,t}^{(i,j)}$ using Eq.~\ref{eq:m}; \\
			\If{$ c=K * T $}{
				Calculate $s_U(o^{(i,j)}_k)$ using Eq.~\ref{eq:UCB}; \\
				Update the search space \{$\Omega^{(i,j)}$\} using Eq.~\ref{eq:min};\\
				$c = 0$; \\
				$K \leftarrow K -1$; \\
			}
		}
	\end{algorithm}
	
	Following UCB in the bandit algorithm, we obtain the initial performance for each operation in every edge. Specifically, we sample one from the $K$ operations in $\Omega^{(i,j)}$ for every edge, then obtain the validation accuracy $a$ which is the initial performance $m_{k,0}^{(i,j)}$ by training adversarially the sampled network for one epoch, and finally assigning this accuracy to all the sampled operations.
	
	By considering the confidence of the $k$th operation with the UCB for every edge, the LCB is calculated by
	 \begin{equation}\label{eq:LCB}
	 \small
	  s_L(o^{(i,j)}_k) = m_{k,t}^{(i,j)} - \sqrt{\frac{2\log N}{n_{k,t}^{(i,j)}}},
	 \end{equation}
	 where $N$ is to the total number of samples, $n_{k,t}^{(i,j)}$ refers to the number of times the $k$th operation of edge $(i,j)$ has been selected, and $t$ is the index of the epoch. The first item in Eq.~\ref{eq:LCB} is the value term which favors the operations that look good historically and the second is the exploration term which allows operations to get an exploration bonus that grows with $\log N$. The selection probability for each operation is defined as
	 
	 \begin{equation}\label{eq:prob}
	 \small
	  p(o^{(i,j)}_k) = \frac{\exp\{-{s_L}(o_k^{(i,j)})\}}{\sum_m \exp\{-{s_L}(o_m^{(i,j)})\}}.
	 \end{equation}
	 The minus sign in Eq.~\ref{eq:prob} means that we prefer to sample operations with a smaller confidence. After sampling one operation for every edge based on $p(o^{(i,j)}_k)$, we obtain the validation accuracy $a$ by training adversarially the sampled network for one epoch, and then update the performance $m_{k,t}^{(i,j)}$ which historically indicates the validation accuracy of all the sampled operations $o^{(i,j)}_k$ as
	\begin{equation}\label{eq:m}
		\small
		m_{k,t}^{(i,j)} = (1 - \lambda) m_{k,t-1}^{(i,j)} + \lambda * a,
	\end{equation}
	where $\lambda$ is a hyper-parameter.
	
	Finally, after $K * T$ samples where $T$ is a hyper-parameter, we calculate the confidence with the UCB according to Eq.~\ref{eq:UCB_o} as
	\begin{equation}\label{eq:UCB}
		\small
		s_U(o^{(i,j)}_k) = m_{k,t}^{(i,j)} + \sqrt{\frac{2\log N}{n_{k,t}^{(i,j)}}}.
	\end{equation}
	The operation with the minimal UCB for every edge is abandoned. This means that the operations that are given more opportunities, but result in poor performance, are removed. With this pruning strategy, the search space is significantly reduced from $|\Omega^{(i,j)}|^{10 \times 6}$ to $(|\Omega^{(i,j)}|-1)^{10 \times 6}$, and the reduced space becomes 
	
	\begin{equation}\label{eq:min}
		\small
		\Omega^{(i,j)} \leftarrow \Omega^{(i,j)} - \{ \mathop{\arg\min}\limits_{o^{(i,j)}_k}{s_U(o^{(i,j)}_k)} \}, \forall(i,j).
	\end{equation}
	The reduction procedure is carried out repeatedly until the optimal structure is obtained where there is only one operation left in each edge. Our anti-bandit search algorithm is summarized in Algorithm \ref{alg:bandit}.

	\textbf{Complexity Analysis.} There are $\mathcal{O}(K^{|\mathcal{E_M}| \times v})$ combinations in the process of finding the optimal architecture in the search space with $v$ kinds of different cells. In contrast, ABanditNAS reduces the search space for every $K * T$ epochs. Therefore, the complexity of the proposed method is
	\begin{equation}
	\small
	\mathcal{O}(T\times \sum_{k=2}^K k) = \mathcal{O}(TK^2).
	\end{equation}

\section{Experiments}\label{img:structure}
	
We demonstrate the robustness of our ABanditNAS on two benchmark datasets (MNIST and CIFAR-10) for the image classification task, and compare ABanditNAS with state-of-the-art robust models.
	
	\subsection{Experiment Protocol}
	
	In our experiments, we search architectures on an over-parameterized network on MNIST and CIFAR-10, and then evaluate the best architecture on corresponding datasets. Unlike previous NAS works \cite{liu2018darts,Xu2019PC,pham2018efficient}, we learn six kinds of cells, instead of two, to increase the diversity of the network. 
	
	{\bfseries Search and Training Settings.} In the search process,  the over-parameterized network is constructed with six cells, where the $2^{nd}$ and $4^{th}$ cells are used to double the channels of the feature maps and halve the height and width of the feature maps, respectively. There are $M=4$ intermediate nodes in each cell. The hyperparameter $T$ which denotes the sampling times is set to $3$, so the total number of epochs is $\sum_{k=2}^Kk*T$. The hyperparameter $\lambda$ is set to $0.7$. The evalution of the hyperparameters is provided in the supplementary file. A large batch size of $512$ is used. And we use an additional regularization cutout \cite{devries2017improved} for CIFAR-10. The initial number of channels is $16$. We employ FGSM adversarial training combined with random initialization and $\epsilon=0.3$ for MNIST, and $\epsilon=0.031$ for CIFAR-10. We use SGD with momentum to optimize the network weights, with an initial learning rate of $0.025$ for MNIST and $0.1$ for CIFAR-10 (annealed down to zero following a cosine schedule), a momentum of 0.9 and a weight decay of $3 \times 10^{-4}$ for MNIST/CIFAR-10.
	
	After search, the six cells are stacked to get the final networks. To adversarially train them, we employ FGSM combined with random initialization and $\epsilon=0.3$ on MNIST, and use PGD-7 with $\epsilon=0.031$ and step size of $0.0078$ on CIFAR-10. Next, we use ABanditNAS-$V$ to represent ABanditNAS with $V$ cells in the training process. The number $V$ can be different from the number $v$. The initial number of channels is $16$ for MNIST, and $48$ for CIFAR-10. We use a batch size of $96$ and an additional regularization cutout \cite{devries2017improved} for CIFAR-10. We employ the SGD optimizer with an initial learning rate of $0.025$ for MNIST and $0.1$ for CIFAR-10 (annealed down to zero following a cosine schedule without restart), a momentum of $0.9$, a weight decay of $3 \times 10^{-4}$, and a gradient clipping at $5$. We train $200$ epochs for MNIST and CIAFR-10.
    
	{\bfseries White-Box vs. Black-Box Attack Settings.} In an adversarial setting, there are two main threat models: white-box attacks where the adversary possesses complete knowledge of the target model, including its parameters, architecture and the training method, and black-box attacks where the adversary feeds perturbed images at test time, which are generated without any knowledge of the target model, and observes the output. We evaluate the robustness of our proposed defense against both settings. The perturbation size $\epsilon$ and step size are the same as those in the adversarial training for both the white-box and black-box attacks. The numbers of iterations for MI-FGSM and BIM are both set to $10$ with a step size and a standard perturbation size the same as those in the white-box attacks. We evaluate ABanditNAS against transfer-based attack where a copy of the victim network is trained with the same training setting. We apply attacks similar to the white-box attacks on the copied network to generate black-box adversarial examples. We also generate adversarial samples using a ResNet-18 model, and feed them to the model obtainedly ABanditNAS.

	\begin{table*}[t]
		\begin{center}
			\resizebox{\textwidth}{12mm}{
				\begin{tabular}{lcccccccc}
					\toprule
					\multirow{2}{*}{\textbf{Architecture}} & {\textbf{Clean}} & {\textbf{FGSM}} & {\textbf{PGD-40}} & {\textbf{PGD-100}} & \textbf{\# Params} & \textbf{Search Cost} & \textbf{Search} \\ 
					& (\%) & (\%) & (\%) & (\%) & \textbf{(M)} & \textbf{(GPU days)} & \textbf{Method} \\ 
					\hline
					LeNet \cite{madry2017towards} & 98.8 & 95.6 &93.2 &91.8 & 3.27 & - & Manual \\ 
					LeNet (Prep. + Adv. train \cite{YangME}) & 97.4 & - & 94.0 &91.8 & 0.06147 & - & Manual \\
					UCBNAS & 99.5 & 98.67 & 96.94 & 95.4 & 0.082 & 0.13 & Bandit \\
					UCBNAS (pruning) & 99.52 & 98.56 & 96.62 & 94.96 & 0.066 & 0.08 & Bandit \\
					\hline
					ABanditNAS-6 & \textbf{99.52} & \textbf{98.94} & \textbf{97.01} & \textbf{95.7} & \textbf{0.089} & \textbf{0.08} & Anti-Bandit \\
					\bottomrule
				\end{tabular}}
			\end{center}
			\caption{Robustness of ABanditNAS under FGSM and PGD attacks on MNIST.}
			\label{tab:mnist_results}
		\end{table*}
		
    \begin{table*}[t]
        \small
        \begin{center}
			\resizebox{\textwidth}{22mm}{
                \begin{tabular}{lc|ccc|ccc}
                    \hline\hline
                    \multicolumn{2}{c}{}&\multicolumn{2}{c}{White-Box}&\multicolumn{1}{c}{}&
                    \multicolumn{3}{c}{Black-Box} \\
                    Structure&Clean &MI-FGSM &BIM &PGD &MI-FGSM &BIM &PGD \\
                    \hline\hline
                    &\multicolumn{7}{c}{MNIST ($\epsilon=0.3$)} \\
                    \hline
					LeNet \cite{madry2017towards} (copy) & 98.8 & - & - & 93.2 & - &- & 96.0 \\ 
                    ABanditNAS-6 (copy) & \textbf{99.52} & 97.41 & 97.63 & \textbf{97.58} & 99.09 & 99.12 & \textbf{99.02} \\
                    \hline\hline
                    &\multicolumn{7}{c}{CIFAR-10 ($\epsilon=0.031$)} \\
                    \hline
					Wide-ResNet \cite{madry2017towards} (copy) & 87.3 & - & - & 50.0 & - & - & 64.2 \\ 
					NASNet \cite{cubuk2017intriguing} (copy) & \textbf{93.2} &- & - & 50.1 & - & - & 75.0 \\ 
                    ABanditNAS-6 (copy) & 87.16 & 48.77 & 47.59 & 50.0 &  74.94 & 75.78 & 76.13 \\
                    ABanditNAS-6 (ResNet-18) & 87.16 & 48.77 & 47.59 & 50.0 & 77.06 & 77.63 & 78.0 \\
                    ABanditNAS-10 (ResNet-18) & 90.64 & \textbf{54.19} & \textbf{55.31} & \textbf{58.74} & \textbf{80.25} & \textbf{80.8} & \textbf{81.26}\\
					\hline\hline
                \end{tabular}}
            \end{center}
            \caption{Robustness of our model in the white-box and black-box settings on MNIST and CIFAR-10. Here $\epsilon$ is the perturbation size. PGD means PGD-40 for MNIST and PGD-7 for CIFAR-10. `copy' means we use a copied network to generate black-box adversarial examples, and `ResNet-18' means using ResNet-18 to generate black-box adversarial examples.}
            \label{tab:w&b_results}
        \end{table*}
	\subsection{Results on Different Datasets}

	\begin{table*}[t]
		\small
		\begin{center}
			\setlength{\tabcolsep}{1mm}{
			\resizebox{\textwidth}{17mm}{
				\begin{tabular}{lccccccccc}
					\toprule
					\multirow{2}{*}{\textbf{Architecture}} & {\textbf{Clean}} & {\textbf{MI-FGSM}} & {\textbf{PGD-7}} & {\textbf{PGD-20}} & \textbf{\# Params} & \textbf{Search Cost} & \textbf{Search} \\ 
					& (\%) & (\%) & (\%) & (\%) & \textbf{(M)} & \textbf{(GPU days)} & \textbf{Method} \\ 
					\hline
					VGG-16 \cite{zhang2019interpreting} & 85.16 & - & 46.04 (PGD-10) & - & - & - & Manual \\
					ResNet \cite{madry2017towards} & 79.4 & - & 47.1 & 43.7 & 0.46 & - & Manual \\
					Wide-ResNet \cite{madry2017towards} & 87.3 & - & 50.0 & 45.8 & 45.9 & - & Manual \\
					NASNet \cite{cubuk2017intriguing} & 93.2 & - & 50.1 & - & - & $\sim7 \times 2000$ & RL \\
					UCBNAS (pruning) & 89.54 & 53.12 & 54.55 & 45.33 & 8.514 & 0.08 & Bandit \\
					\hline
					ABanditNAS-6 & 87.16 & 48.77 & 50.0 & 45.9 & \textbf{2.892} & 0.08 & Anti-Bandit \\
					ABanditNAS-6 (larger) & 87.31 & 52.01 & 51.24 & 45.79 & 12.467 & 0.08 & Anti-Bandit \\
					ABanditNAS-10 & 90.64 & \textbf{54.19} & \textbf{58.74} & \textbf{50.51} & 5.188 & 0.08 & Anti-Bandit \\
					\bottomrule
				\end{tabular}}}
			\end{center}
			\caption{Validation accuracy and robustness of various models trained on CIFAR-10. Note that the search cost of NASNet which is unknown is estimated based on \cite{cubuk2017intriguing}. `PGD-10' means the result of VGG-16 is under PGD-10 attack which comes from \cite{zhang2019interpreting}.}
			\label{tab:cifar10_results}
		\end{table*}
	
\begin{figure}[thbp]
		\centering
		\subfigure[MNIST]{
			\includegraphics[scale=.4]{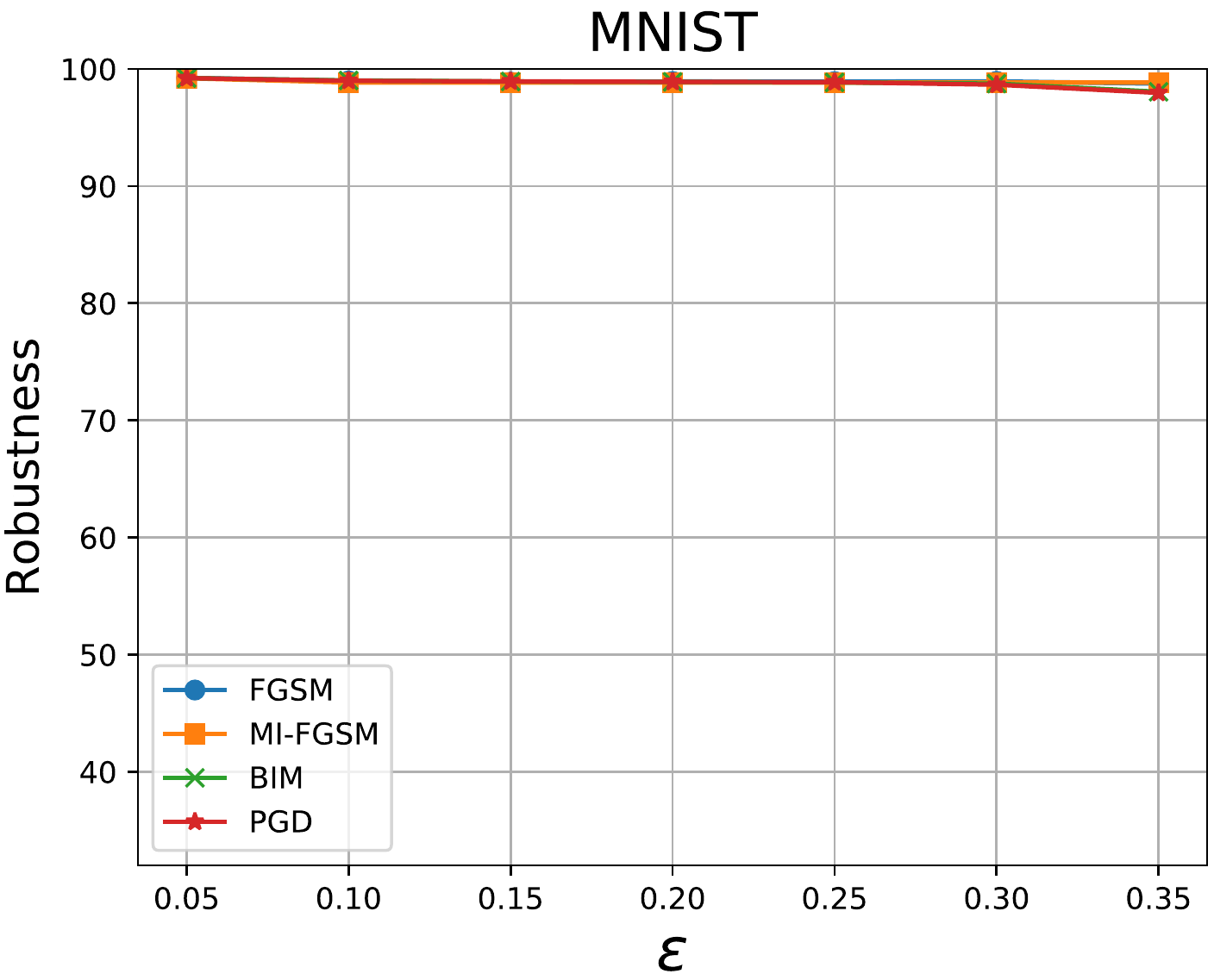}
			\label{fig:eps_mnist}
		}
		\subfigure[CIFAR-10]{
			\includegraphics[scale=.4]{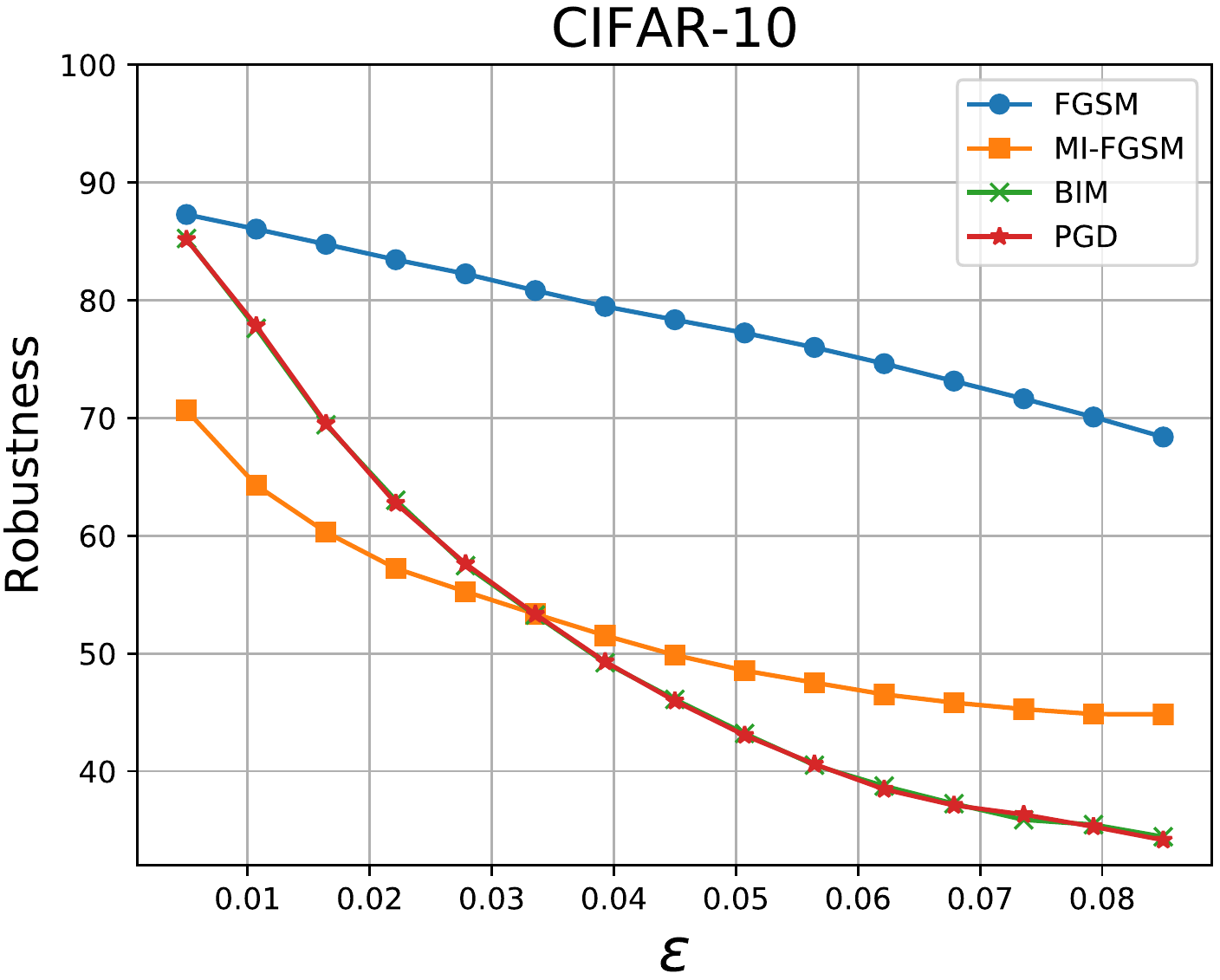}
			\label{fig:eps_cifar10}
		}
		\caption{Robustness of ABanditNAS against different white-box attacks for various perturbation budgets.}	
	\end{figure}

	{\bfseries MNIST.} Owing to the search space reduction by anti-bandit, the entire search process only requires $1.93$ hours on a single NVIDIA Titan V GPU. For MNIST, the structure searched by ABanditNAS is directly used for training. We evaluate the trained network by $40$ and $100$ attack steps, and compare our method with LeNet \cite{madry2017towards} and MeNet \cite{YangME} in Table \ref{tab:mnist_results}. From these results, we can see that ABanditNAS using FGSM adversarial training with random initialization is more robust than LeNet with PGD-40 adversarial training, no matter which attack is used. Although MeNet uses matrix estimation (ME) as preprocessing to destroy the adversarial structure of the noise, our method still performs better. In addition, our method has the best performance ($99.52\%$) on the clean images with a strong robustness. For the black-box attacks, Table \ref{tab:w&b_results} shows that they barely affect the structures searched by ABanditNAS compared with other models, either manually designed or searched by NAS. As illustrated in Fig.~\ref{fig:eps_mnist}, with the increase of the perturbation size, our network's performance does not drop significantly, showing the robustness of our method.

    We also apply the conventional bandit which samples operations based on UCB to search the network, leading to UCBNAS. The main differences between UCBNAS and ABanditNAS lie in that UCBNAS only uses UCB as an evaluation measure to select an operation, and there is no operation pruning involved.  Compared with UCBNAS, ABanditNAS can get better performance and use less search time under adversarial attacks as shown in Table \ref{tab:mnist_results}. Also, to further demonstrate the effectiveness of our ABanditNAS, we use UCBNAS with pruning to search for a robust model, which not only uses UCB to select an operation, but also prune operation of less potential. Although UCBNAS (pruning) is as fast as ABanditNAS, it has  worse performance than ABanditNAS beceuse of unfair competitions between operations before pruning.

	{\bfseries CIFAR-10.} The results for different architectures on CIFAR-10 are summarized in Table \ref{tab:cifar10_results}. We use one Titan V GPU to search, and the batch size is $512$. The entire search process takes about $1.94$ hours. We consider $V = 6$ and $V = 10$ cells for training. In addition, we also train a larger network variant with $100$ initial channels for $V = 6$. Compared with Wide-ResNet, ABanditNAS-10 achieves not only a better performance ($50.0$\% vs. $58.74$\%) in PGD-7, but also fewer parameters ($45.9$M vs. $5.188$M). Although the result of VGG-16 is under PGD-10, ABanditNAS-10 achieves a better performance under more serious attack PGD-20 ($46.04$\% vs. $50.51$\%). When compared with NASNet\footnote{Results are from \cite{cubuk2017intriguing}.} which has a better performance on clean images, our method obtains better performance on adversarial examples with a much faster search speed ($\sim 7 \times 2000$ vs. $0.08$). Note that the results in Table \ref{tab:cifar10_results} are the best we got, which are unstable and need more trials to get the results. Table \ref{tab:w&b_results} shows the black-box attacks barely affect the networks obtained by ABanditNAS, much less than those by other methods. In addition, Fig.~\ref{fig:eps_cifar10} illustrates ABanditNAS is still robust when the disturbance increases.

	\begin{figure}[t]
		\centering
		\subfigure[First Cell]{
			\includegraphics[width=3.59cm]{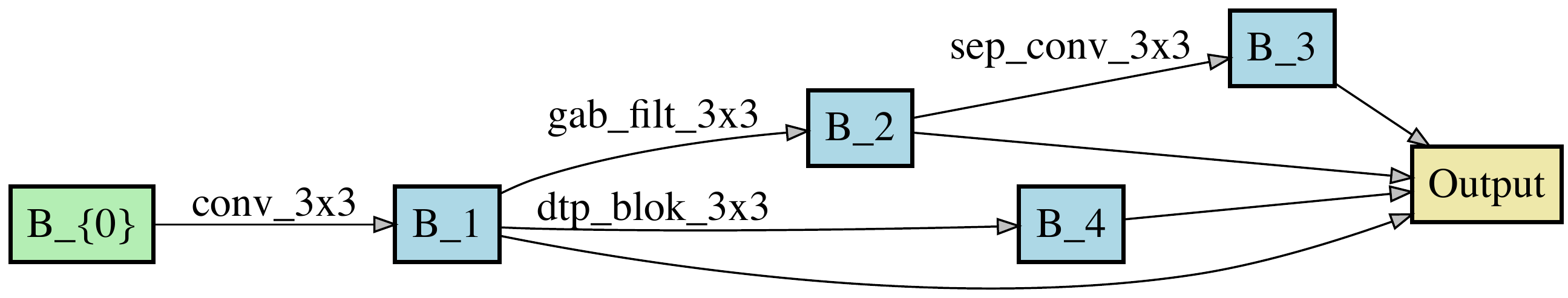}
		}
		\subfigure[Second Cell]{
			\includegraphics[width=3.59cm]{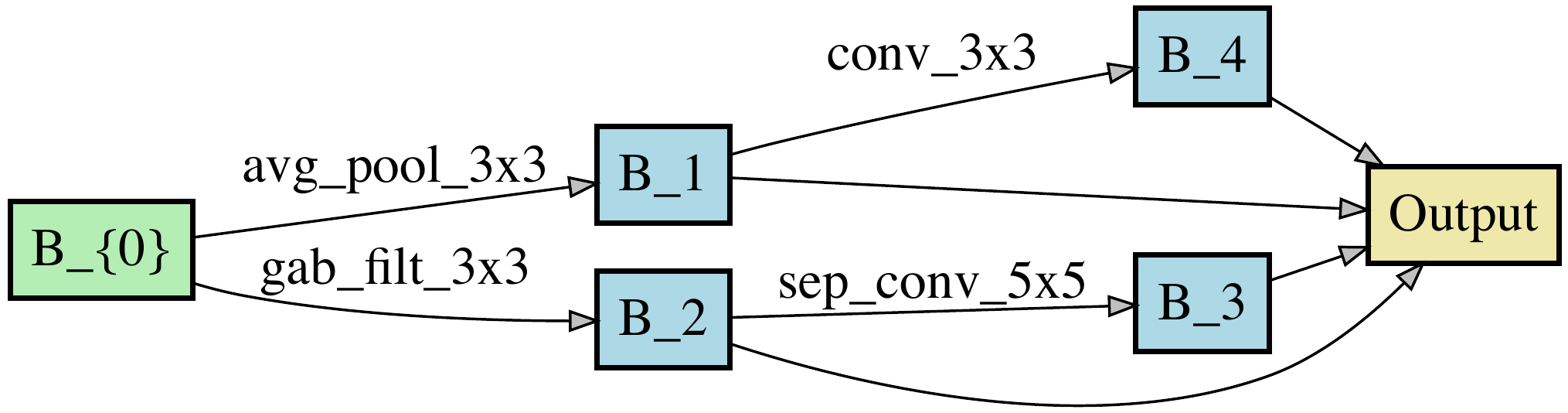}
		}
		\subfigure[Third Cell]{
			\includegraphics[width=3.35cm]{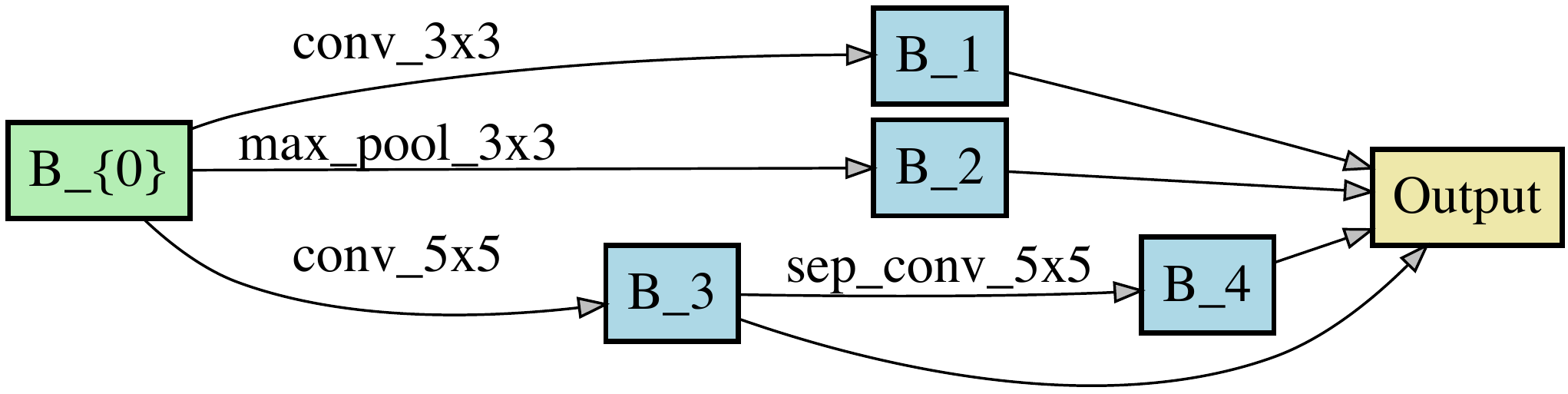}
		}
		\quad
		\subfigure[Fourth Cell]{
			\includegraphics[width=3.59cm]{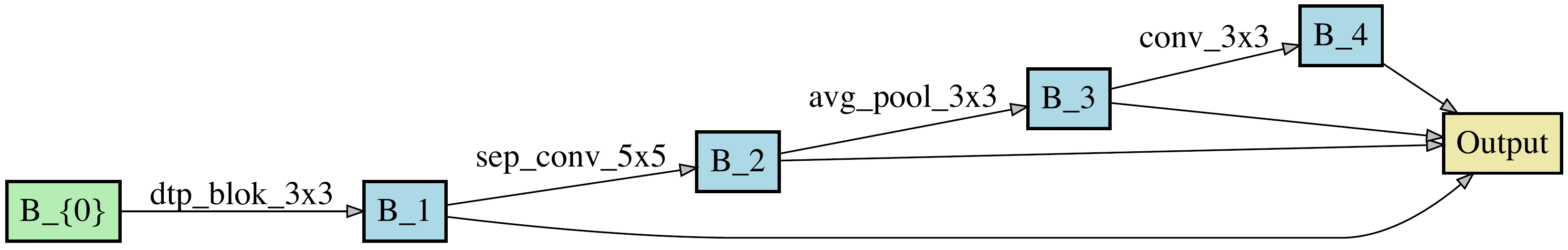}
		}
		\subfigure[Fifth Cell]{
			\includegraphics[width=3.59cm]{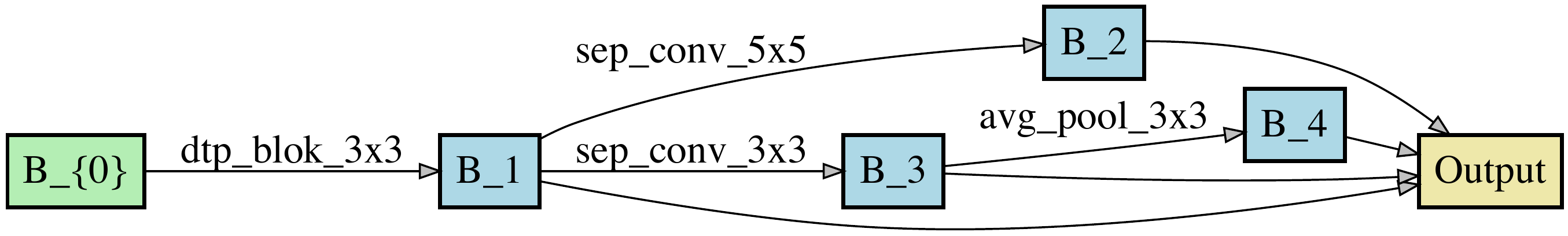}
		}
		\subfigure[Sixth Cell]{
			\includegraphics[width=3.59cm]{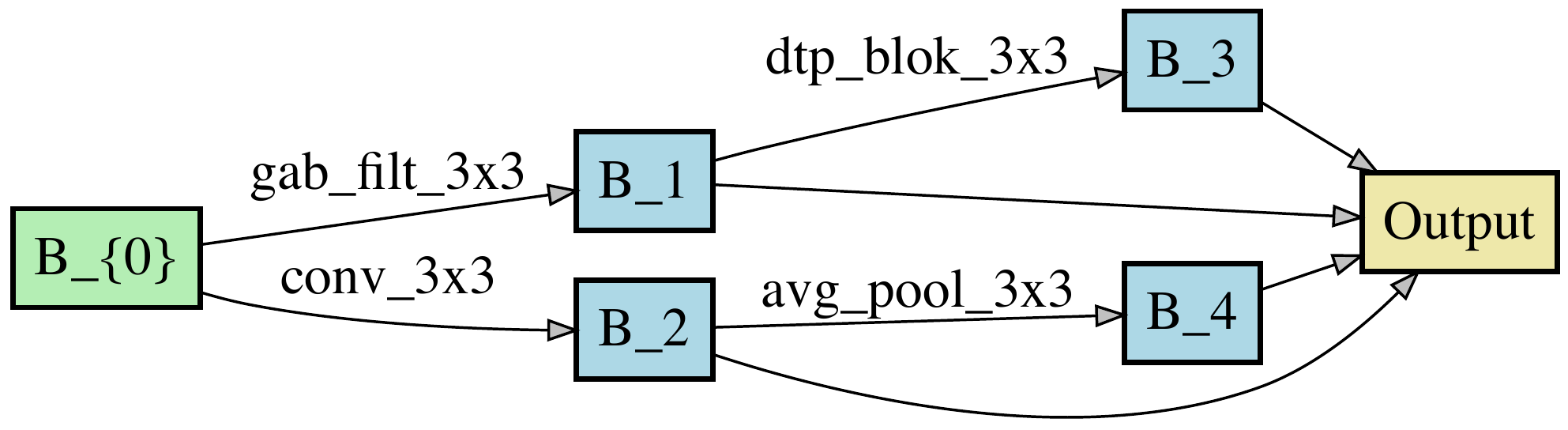}
		}
		\caption{Detailed structures of the best cells discovered on CIFAR-10 using FGSM with random initialization.}
		\label{fig:structures}
	\end{figure}
	
	For the structure searched by ABanditNAS on CIFAR-10, we find that the robust structure prefers pooling operations, Gabor filters and denosing blocks (Fig.~\ref{fig:structures}). The reasons lie in that the pooling can enhance the nonlinear modeling capacity, Gabor filters can extract robust features, and the denosing block and mean pooling act as smoothing filters for denosing. Gabor filters and denosing blocks are usually set in the front of cell by ABanditNAS to denoise feature encoded by the previous cell. The setting is consistent with  \cite{xie2019feature}, which demonstrates the rationality of ABanditNAS.

	\subsection{Ablation Study}
	
	The performances of the structures searched by ABanditNAS with different values of the $\lambda$ are used to find the best $\lambda$. We train the structures under the same setting. 
 
\begin{figure}[h]
		\centering
			\includegraphics[scale=.45]{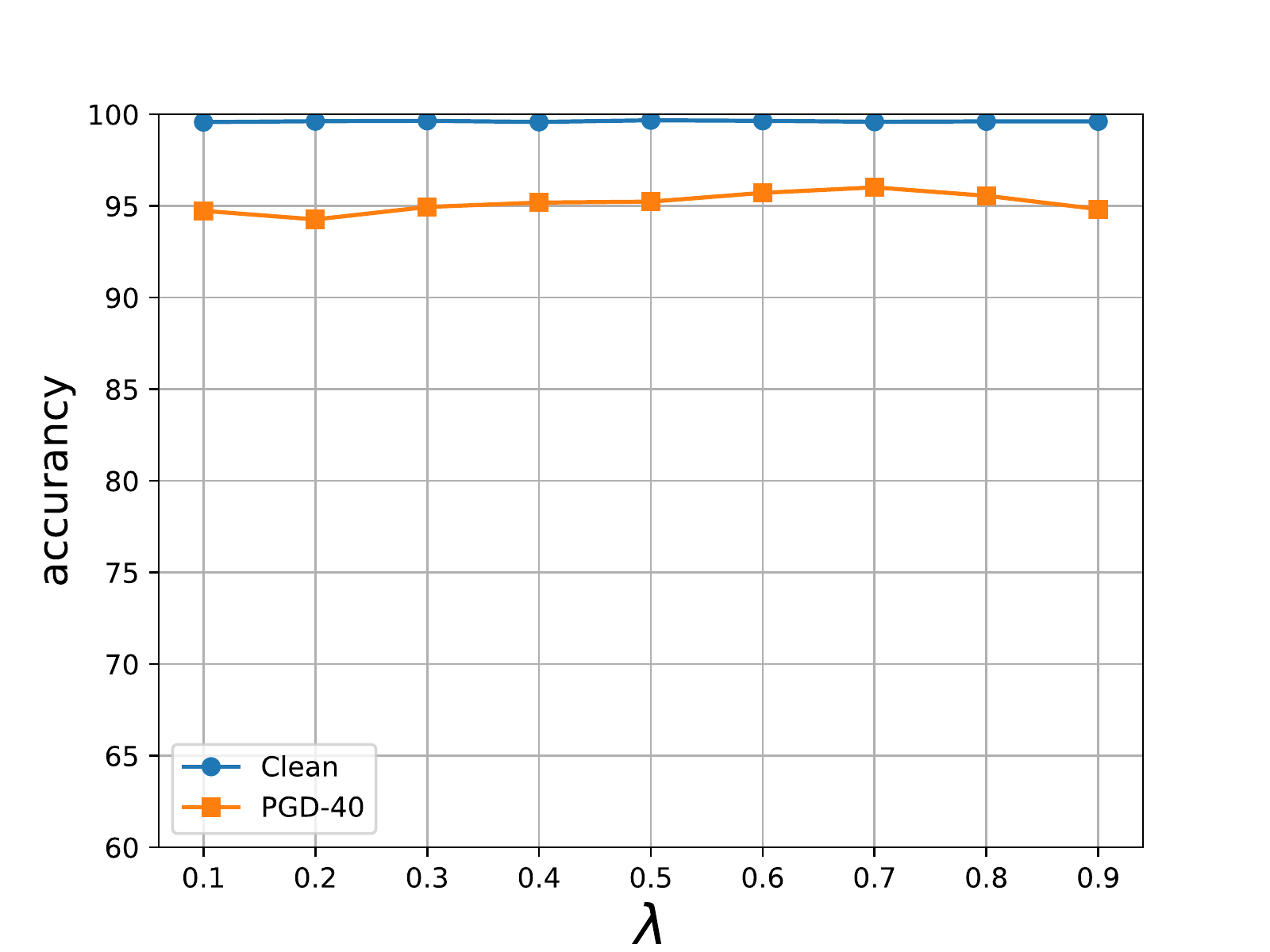}
			\label{fig:lambda}
		\caption{The performances of the structures searched by ABanditNAS with different values of the hyperparameters $T$ and $\lambda$.}	
	\end{figure}

    \textbf{Effect on the hyperparameter $\lambda$:}
    The hyperparameter $\lambda$ is used to balance the performance between the past and the current. Different values of $\lambda$ result in similar search costs. From Fig.~\ref{fig:lambda}, we can see that when $\lambda=0.7$, ABanditNAS is most robust.
    
\section{Conclusion}

We have proposed an ABanditNAS approach to  design robust structures to defend adversarial attacks. To solve the challenging search problem caused by the  complicated huge search space and the adversarial training process, we have introduced an anti-bandit algorithm to improve the search efficiency. We have investigated the relationship between our  strategy and potential operations based on both  lower and  upper bounds. Extensive experiments have demonstrated that the proposed ABanditNAS is much faster than other state-of-the-art NAS methods with a better performance in accuracy. Under adversarial attacks, ABanditNAS achieves much better performance than other  methods.

\section*{Acknowledgement}
Baochang Zhang is also with  Shenzhen Academy of Aerospace Technology, Shenzhen, China, and he is the corresponding author.  He is  in part Supported by National Natural Science Foundation of China under Grant 61672079, Shenzhen Science and Technology Program (No.KQTD2016112515134654)

\clearpage
%
%
\bibliographystyle{splncs04}
\bibliography{egbib}

\begin{thebibliography}{10}
\providecommand{\url}[1]{\texttt{#1}}
\providecommand{\urlprefix}{URL }
\providecommand{\doi}[1]{https://doi.org/#1}

\bibitem{athalye2018obfuscated}
Athalye, A., Carlini, N., Wagner, D.: Obfuscated gradients give a false sense
  of security: Circumventing defenses to adversarial examples. In: ICML (2018)

\bibitem{auer2002finite}
Auer, P., Cesa-Bianchi, N., Fischer, P.: Finite-time analysis of the multiarmed
  bandit problem. Machine learning  (2002)

\bibitem{BenderKZVL18}
Bender, G., Kindermans, P.J., Zoph, B., Vasudevan, V., Le, Q.V.: Understanding
  and simplifying one-shot architecture search. In: ICML (2018)

\bibitem{bengio2017deep}
Bengio, Y., Goodfellow, I., Courville, A.: Deep learning. Citeseer (2017)

\bibitem{buades2005non}
Buades, A., Coll, B., Morel, J.: A non-local algorithm for image denoising. In:
  CVPR (2005)

\bibitem{carlini2017towards}
Carlini, N., Wagner, D.: Towards evaluating the robustness of neural networks.
  In: 2017 IEEE Symposium on Security and Privacy (2017)

\bibitem{chen2020bnas}
Chen, H., Zhuo, L., Zhang, B., Zheng, X., Liu, J., Doermann, D., Ji, R.:
  Binarized neural architecture search. AAAI  (2020)

\bibitem{cubuk2017intriguing}
Cubuk, E.D., Zoph, B., Schoenholz, S.S., Le, Q.V.: Intriguing properties of
  adversarial examples. In: ICLR (2017)

\bibitem{das2017keeping}
Das, N., Shanbhogue, M., Chen, S., Hohman, F., Chen, L., Kounavis, M.E., Chau,
  D.H.: Keeping the bad guys out: Protecting and vaccinating deep learning with
  jpeg compression. arXiv  (2017)

\bibitem{devries2017improved}
DeVries, T., Taylor, G.W.: Improved regularization of convolutional neural
  networks with cutout. arXiv  (2017)

\bibitem{dong2018boosting}
Dong, Y., Liao, F., Pang, T., Su, H., Zhu, J., Hu, X., Li, J.: Boosting
  adversarial attacks with momentum. In: CVPR (2018)

\bibitem{vargas2019evolving}
D.V.Vargas, S.Kotyan: Evolving robust neural architectures to defend from
  adversarial attacks. arXiv  (2019)

\bibitem{dziugaite2016study}
Dziugaite, G.K., Ghahramani, Z., Roy, D.M.: A study of the effect of jpg
  compression on adversarial images. arXiv  (2016)

\bibitem{even2006qlearning}
Even-Dar, E., Mannor, S., Mansour, Y.: Action elimination and stopping
  conditions for the multi-armed bandit and reinforcement learning problems.
  Journal of Machine Learning Research  (2006)

\bibitem{gabor1946electrical}
Gabor, D.: Electrical engineers part iii: Radio and communication engineering,
  j. Journal of the Institution of Electrical Engineers - Part III: Radio and
  Communication Engineering 1945-1948  (1946)

\bibitem{gabor1946theory}
Gabor, D.: Theory of communication. part 1: The analysis of information.
  Journal of the Institution of Electrical Engineers-Part III: Radio and
  Communication Engineering  (1946)

\bibitem{goodfellow2014explaining}
Goodfellow, I.J., Shlens, J., Szegedy, C.: Explaining and harnessing
  adversarial examples. arXiv  (2014)

\bibitem{gupta2019ciidefence}
Gupta, P., Rahtu, E.: Ciidefence: Defeating adversarial attacks by fusing
  class-specific image inpainting and image denoising. In: ICCV (2019)

\bibitem{he2016deep}
He, K., Zhang, X., Ren, S., Sun, J.: Deep residual learning for image
  recognition. In: CVPR (2016)

\bibitem{ilyas2018prior}
Ilyas, A., Engstrom, L., Madry, A.: Prior convictions: Black-box adversarial
  attacks with bandits and priors. In: ICLR (2018)

\bibitem{lev2006mc}
Kocsis, L., Szepesvari, C.: Bandit based monte-carlo planning. Proceedings of
  the 17th European conference on Machine Learning  (2006)

\bibitem{kurakin2016adversarial}
Kurakin, A., Goodfellow, I., Bengio, S.: Adversarial examples in the physical
  world. In: ICLR (2016)

\bibitem{liao2018defense}
Liao, F., Liang, M., Dong, Y., Pang, T., Hu, X., Zhu, J.: Defense against
  adversarial attacks using high-level representation guided denoiser. In: CVPR
  (2018)

\bibitem{liu2018darts}
Liu, H., Simonyan, K., Yang, Y.: Darts: Differentiable architecture search. In:
  ICLR (2018)

\bibitem{liu2016delving}
Liu, Y., Chen, X., Liu, C., Song, D.: Delving into transferable adversarial
  examples and black-box attacks. In: ICLR (2016)

\bibitem{long2015fully}
Long, J., Shelhamer, E., Darrell, T.: Fully convolutional networks for semantic
  segmentation. In: CVPR (2015)

\bibitem{madry2017towards}
Madry, A., Makelov, A., Schmidt, L., Tsipras, D., Vladu, A.: Towards deep
  learning models resistant to adversarial attacks. In: ICLR (2017)

\bibitem{guo2019meets}
M.Guo, Y.Yang, R.Xu, Z.Liu: When nas meets robustness: In search of robust
  architectures against adversarial attacks. CVPR  (2020)

\bibitem{na2017cascade}
Na, T., Ko, J.H., Mukhopadhyay, S.: Cascade adversarial machine learning
  regularized with a unified embedding. In: ICLR (2017)

\bibitem{dong2019neural}
N.Dong, M.Xu, X.Liang, Y.Jiang, W.Dai, E.Xing: Neural architecture search for
  adversarial medical image segmentation. In: MICCAI (2019)

\bibitem{osadchy2017no}
Osadchy, M., Hernandez-Castro, J., Gibson, S., Dunkelman, O., P{\'e}rez-Cabo,
  D.: No bot expects the deepcaptcha! introducing immutable adversarial
  examples, with applications to captcha generation. IEEE Transactions on
  Information Forensics and Security  (2017)

\bibitem{pham2018efficient}
Pham, H., Guan, M.Y., Zoph, B., Le, Q.V., Dean, J.: Efficient neural
  architecture search via parameter sharing. In: ICML (2018)

\bibitem{RobustGabor}
Pérez, J.C., Alfarra, M., Jeanneret, G., Bibi, A., Thabet, A.K., Ghanem, B.,
  Arbeláez, P.: Robust gabor networks. arXiv  (2019)

\bibitem{shafahi2019adversarial}
Shafahi, A., Najib, M., A, A.G., Xu, Z., Dickerson, J., Studer, C., Davis,
  L.S., Taylor, G., Goldstein, T.: Adversarial training for free! In: NIPS
  (2019)

\bibitem{alphago}
Silver, D., S., J., S., K., etc., I.A.: Mastering the game of go without human
  knowledge. In: Nature (2017)

\bibitem{samangouei2018defensegan}
S.Pouya, K.Maya, C.Rama: Defense-{GAN}: Protecting classifiers against
  adversarial attacks using generative models. In: ICLR (2018)

\bibitem{szegedy2015going}
Szegedy, C., Liu, W., Jia, Y., Sermanet, P., Reed, S., Anguelov, D., Erhan, D.,
  Vanhoucke, V., Rabinovich, A.: Going deeper with convolutions. In: CVPR
  (2015)

\bibitem{szegedy2013intriguing}
Szegedy, C., Zaremba, W., Sutskever, I., Bruna, J., Erhan, D., Goodfellow, I.,
  Fergus, R.: Intriguing properties of neural networks. In: ICLR (2013)

\bibitem{wang2019bilateral}
Wang, J., Zhang, H.: Bilateral adversarial training: Towards fast training of
  more robust models against adversarial attacks. In: ICCV (2019)

\bibitem{Wong2020Fast}
Wong, E., Rice, L., Kolter, J.Z.: Fast is better than free: Revisiting
  adversarial training. In: ICLR (2020)

\bibitem{xie2019feature}
Xie, C., Wu, Y., Maaten, L.V.D., Yuille, A.L., He, K.: Feature denoising for
  improving adversarial robustness. In: CVPR (2019)

\bibitem{Xu2019PC}
Xu, Y., Xie, L., Zhang, X., Chen, X., Qi, G., Tian, Q., Xiong, H.: Pc-darts:
  Partial channel connections for memory-efficient differentiable architecture
  search. In: ICLR (2019)

\bibitem{YangME}
Yang, Y., Zhang, G., Katabi, D., Xu, Z.: Me-net: Towards effective adversarial
  robustness with matrix estimation. In: ICML (2019)

\bibitem{ying2019bench}
Ying, C., Klein, A., Real, E., Christiansen, E., Murphy, K., Hutter, F.:
  Nas-bench-101: Towards reproducible neural architecture search. In: ICML
  (2019)

\bibitem{zhang2019interpreting}
Zhang, C., Liu, A., Liu, X., Xu, Y., Yu, H., Ma, Y., Li, T.: Interpreting and
  improving adversarial robustness with neuron sensitivity. arXiv  (2019)

\bibitem{zheng2019dynamic}
Zheng, X., Ji, R., Tang, L., Wan, Y., Zhang, B., Wu, Y., Wu, Y., Shao, L.:
  Dynamic distribution pruning for efficient network architecture search. arXiv
   (2019)

\bibitem{zhuo2020cp}
Zhuo, L., Zhang, B., Chen, H., Yang, L., Chen, C., Zhu, Y., Doermann, D.:
  Cp-nas: Child-parent neural achitecture search for 1-bit cnns. IJCAI  (2020)

\bibitem{zoph2016neural}
Zoph, B., Le, Q.V.: Neural architecture search with reinforcement learning. In:
  ICLR (2016)

\bibitem{zoph2018learning}
Zoph, B., Vasudevan, V., Shlens, J., Le, Q.V.: Learning transferable
  architectures for scalable image recognition. In: CVPR (2018)

\end{thebibliography}
\end{document}